\documentclass[letterpaper]{article}
\usepackage{apacite}  

\usepackage{aaai18}  
\usepackage{times}  
\usepackage{helvet}  
\usepackage{courier}  
\usepackage{url}  
\usepackage{graphicx}  
\usepackage[utf8]{inputenc} 
\usepackage[T1]{fontenc}    
\usepackage{hyperref}       
\usepackage{url}            
\usepackage{booktabs}       
\usepackage{amsfonts}       
\usepackage{nicefrac}       
\usepackage{microtype}      
\usepackage{color}
\usepackage{booktabs}

\usepackage{balance}

\usepackage{dblfloatfix}    
\usepackage{mathtools}
\usepackage{amssymb}
\usepackage{amsmath}
\usepackage{tabu}

\newcommand{\AMI}{\text{AMI}}

\newcommand{\brock}[1]{\mathopen{}\mathclose\bgroup\left[#1\aftergroup\egroup\right]}

\newcommand{\expec}[2]{\mathbb{E}_{#2} \brock{#1}
}
\newcommand{\estexpec}[2]{\hat{\mathbb{E}}_{#2} \brock{#1}
}

\newcommand{\numrollouts}{1,000 }

\title{Burn-In Demonstrations for Multi-Modal Imitation Learning}

\author{
  Alex Kuefler \\ 
  Symbolic Systems Program\\
  Stanford University\\
  Stanford, CA 94305 \\
  \texttt{akuefler@stanford.edu} \\
  \And
  Mykel J. Kochenderfer \\ 
  Department of Aeronautics \& Astronautics\\
  Stanford University\\
  Stanford, CA 94305 \\
  \texttt{mykel@stanford.edu} \\
}

\begin{document}

\maketitle

\begin{abstract}
Recent work on imitation learning has generated policies that reproduce expert behavior from multi-modal data. However, past approaches have focused only on recreating a small number of distinct, expert maneuvers, or have relied on supervised learning techniques that produce unstable policies. This work extends InfoGAIL, an algorithm for multi-modal imitation learning, to reproduce behavior over an extended period of time. Our approach involves reformulating the typical imitation learning setting to include ``burn-in demonstrations'' upon which policies are conditioned at test time. We demonstrate that our approach outperforms standard InfoGAIL in maximizing the mutual information between predicted and unseen style labels in road scene simulations, and we show that our method leads to policies that imitate expert autonomous driving systems over long time horizons.
\end{abstract}


\section{Introduction}

Modeling human behavior is necessary for developing and validating autonomous systems. In the context of autonomous driving, modeling drivers is challenging because there is significant variability in driving style and behavior. Latent factors, such as a person's degree of attentiveness or their willingness to take risks may influence the type of driving behavior they demonstrate. As a result, a distribution of expert demonstrations of some sequential decision making task may have multiple modes, resulting from factors that are difficult to measure.

One line of research attempts to discover latent factors underlying expert demonstrations using fully differentiable models trained with stochastic gradient variational Bayes \cite{kingma2013auto,watter2015embed}. In robotics, variational autoencoders (VAE) have been used to discover latent embeddings of human demonstrations, allowing classical controllers to act in feature spaces that obey desirable properties \cite{watter2015embed}. VAEs have also been used to learn shared embedding spaces for different \emph{sensor} modalities, allowing a single model to reconstruct, for example, the motion of a stroke from an image of a handwritten digit \cite{yin2017associate}. More recently, this model family has been applied to discover different \emph{actuation} modalities, as in the case of demonstrations that share the same observation space, but were sampled from experts who obey different policies. In the context of autonomous driving, driver modeling is treated as a conditional density estimation problem, where the model is trained by conditioning on driver observations and predicting actions (e.g., acceleration and turnrate) from the expert demonstrations alone. Such models can be fit without gathering new data in simulation, and can thus discover latent factors in expert demonstrations directly \cite{morton2017simultaneous}. However, policies trained with supervised learning are sensitive to minor prediction errors, making this approach impractical for many sequential decision making problems \cite{ross2010efficient}.

Alternatively, methods based on Generative Adversarial Imitation Learning (GAIL) combine supervised and reinforcement learning by conducting rollouts in a simulation environment \cite{ho16}. Human demonstrations and policy rollouts can then be compared by a critic, which is trained to provide high reward when the policy's behavior becomes indistinguishable from those of experts. Information Maximizing GAIL (InfoGAIL), in particular, addresses the problem of learning policies from multi-modal demonstrations, and has been used to produce driver models that can give rise to different passing and turning behaviors \cite{li2017inferring}. However, InfoGAIL and related techniques \cite{hausman2017multi} involve sampling a latent code at the beginning of each trial. If the simulated ego-vehicle is initialized with the velocity and heading of a real driver, the random sampling of latent codes can not ensure consistency between the policy's subsequent actions and the driver's true style. This shortcoming limits the applicability of InfoGAIL to modeling real highway scenes, where ego vehicles are sampled from playbacks of recorded human data \cite{kuefler2017imitating}.

We introduce \emph{Burn-InfoGAIL}, an imitation learning technique that addresses this limitation by drawing latent codes directly from a learned, inference distribution \cite{zhao2017towards}. Like recent work on one-shot \cite{duan2017one} and diverse imitation learning \cite{wang2017robust}, our models not only learn from a set of demonstrations, but also condition upon specific reference demonstrations at the beginning of each rollout in a simulated environment. However, Burn-InfoGAIL assumes a new task formulation, motivated by simulated driving. In this setting, a policy must take over from the point at which a specific expert demonstration ends, such as when steering is engaged in an autonomous car. We refer to the partial, expert trajectory as a \emph{burn-in} demonstration, upon which the learned inference model must be conditioned in order to draw latent codes. This work demonstrates that Burn-InfoGAIL is able to achieve greater adjusted mutual information (AMI) with true driver styles than standard InfoGAIL or a variational autoencoder (VAE) baseline. Furthermore, we show that driving trajectories produced by Burn-InfoGAIL deviate less from expert demonstrations than GAIL, InfoGAIL, or supervised learning techniques.

\begin{figure}[!t]
\begin{center}
\includegraphics[scale=.32]{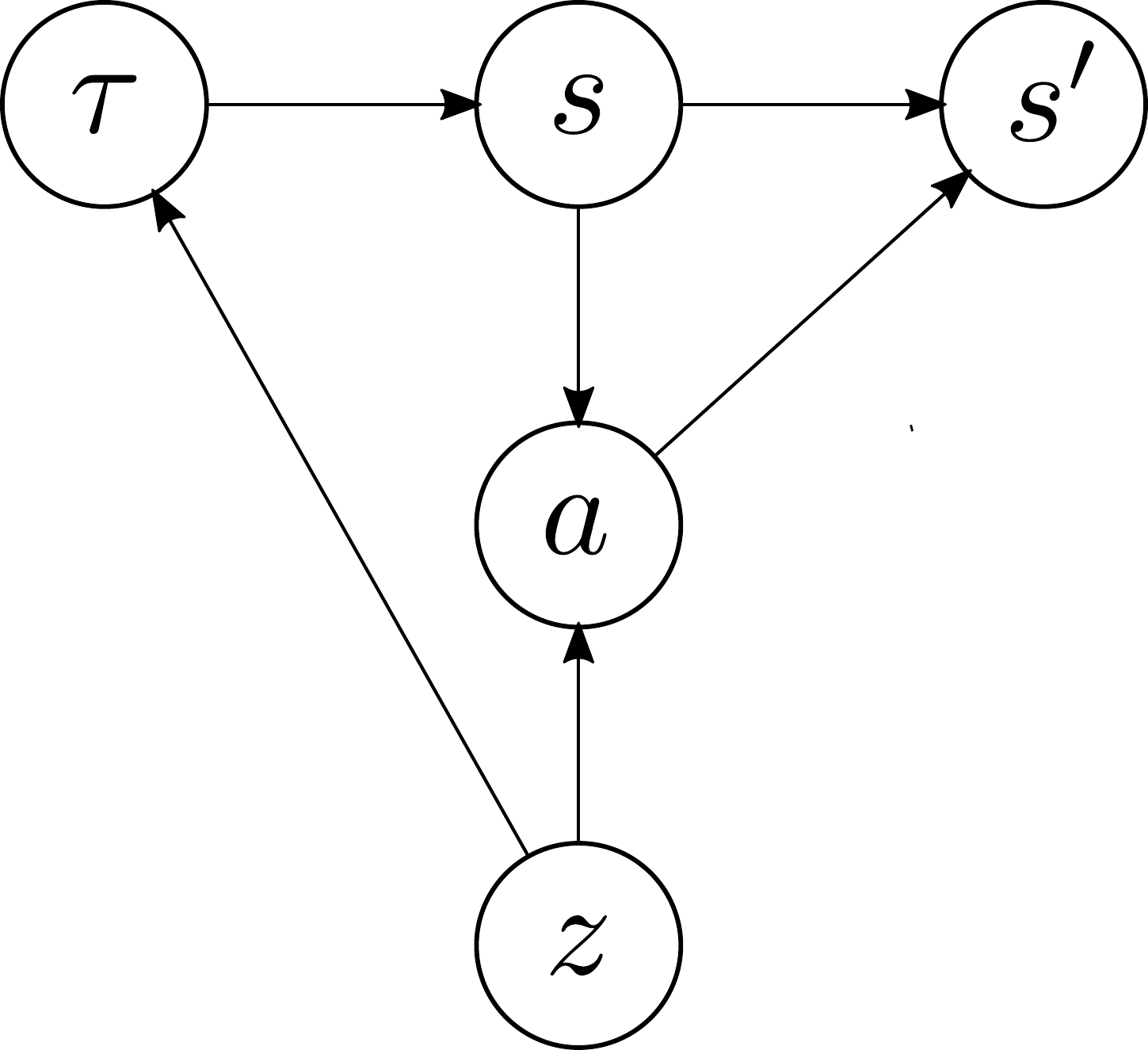}
\caption{Dynamic Bayesian network model of driver style. A latent factor $z$ determines the underlying style of driver behavior. Vehicles progress according to an expert rollout $\tau$, which is a sequence of states and actions carried out by a human driver or hand-crafted controller. The learned policy is conditioned on this history to select an action $a$, which positions the model in a subsequent state $s'$ as determined by the dynamics of the environment.}
\label{fig:DBN}
\end{center}
\end{figure}



\begin{figure*}[!t]
\begin{center}
\includegraphics[scale=.55]{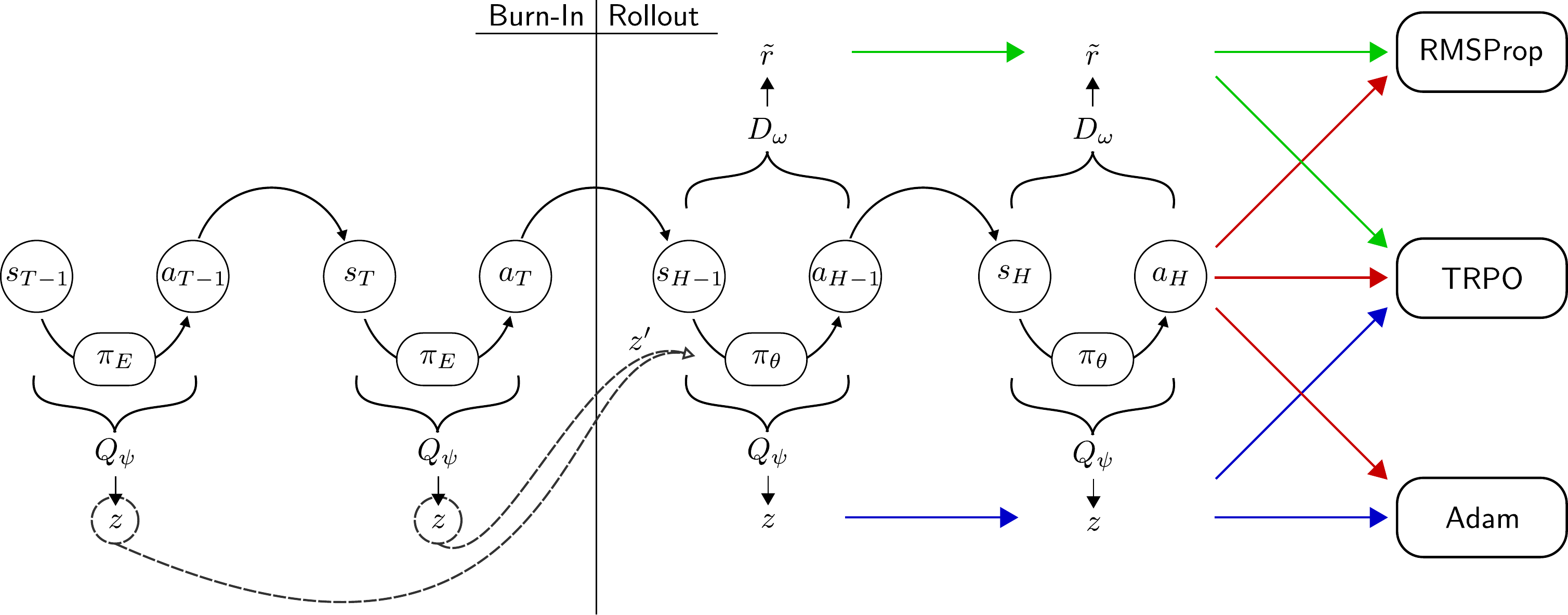}
\caption{Diagram of Burn-InfoGAIL. The expert $\pi_E$ selects actions during the burn-in demonstration, whereas learned $\pi_\theta$ selects actions during the rollout. Dashed lines represent the majority vote taken over predicted latent codes $z$ to produce the initial $z'$ for the rollout. Red arrows represent the contribution of the state-action pairs to the RMSProp, TRPO, and Adam optimizers. Blue arrows represent the contribution of $z$, and green arrows represent the contribution of critic outputs $\tilde{r}$.}
\label{fig:burn-infogail}
\end{center}
\end{figure*}

\section{Problem Formulation}

We adopt a dynamic Bayesian network model of driver style \cite{morton2017simultaneous}. Each vehicle is characterized by a unique \emph{style} variable $z$, which influences the action $a_t$ taken in response to an observation $s_t$ seen at time $t$. In this work, we assume a vehicle's trajectory through this environment proceeds in two stages. First, actions are chosen according to an expert policy $\pi_{E}$ obeying style $z$ for a burn-in demonstration lasting $T$ time steps beginning by first observing $s_0$. Starting with $s_{T+1}$, actions are then sampled until early termination or time horizon $H$ according to a learned policy $\pi_{\theta}$, parameterized by $\theta$. We will use $\tau = (s_0, a_0, ..., s_T, a_T)$ to denote the sequence of observations and actions occurring during the burn-in demonstration. The generative process that gives rise to our data (shown in Figure~\ref{fig:DBN}) factorizes according to:
\begin{align}
\label{eq:dbn1}
p(s,a,z,\tau) = p(z)p(\tau \mid z)p(s \mid \tau)p(a \mid s, z) \\
\label{eq:dbn2}
= p(\tau)p(s \mid \tau)p(z \mid \tau)p(a \mid s, z)
\end{align}
where $s = s_{T+1}$, and the factors $p(a \mid s, z)$ and $p(\tau)$ may be interpreted as $\pi_{\theta}$ and a distribution over expert trajectories respectively. The factor $p(s \mid \tau)$ corresponds to the transition dynamics of the environment, which leaves $p(z \mid \tau)$ to be estimated from data. In our setting, the actions $a$ are two dimensional vectors encoding the acceleration and turn-rate of the ego-vehicle. The observation $s$ consists of both hand-selected and low-level features, described in the implementation section.

\section{Approach}

We propose a new variation to GAIL, which discovers latent factors in expert demonstrations while learning different driving policies. Unlike past work, we assume a setting in which our policy not only learns from demonstrations, but conditions on individual trajectories, continuing from where expert demonstrations stopped. This section describes the objectives we wish to optimize in order to discover both latent intentions and stable policies.

\subsection{Imitation Learning}

In the imitation learning setting, we wish to train a policy $\pi_{\theta}$ that captures behavior similar to those of an expert policy $\pi_{E}$. Because the reward optimized by $\pi_{E}$ is unknown, GAIL \cite{ho16} introduces a discriminator $D_{\omega}$, parameterized by $\omega$, that can help $\pi_{\theta}$ improve by distinguishing expert from non-expert actions. GAIL minimizes with respect to $\theta$ and maximizes with respect to $\omega$ the objective:
\begin{align}
\label{eq:imit}
\begin{split}
V(\theta, \omega) = \expec{\log D_{\omega}(s,a)}{a \sim \pi_{E}(\cdot \mid s)} +\\\expec{\log(1 - D_{\omega}(s,a))}{a \sim \pi_{\theta}(\cdot \mid s)} 
\end{split}
\end{align}
Recent variants of GAIL \cite{li2017inferring} replace the discriminator with a \emph{critic}, which outputs a real-valued score rather than a probability. We adopt this formulation and train $D_{\omega}$ to minimize the Wasserstein objective,
\begin{align}
\label{eq:wasserstein}
\begin{split}
W(\theta, \omega) = \expec{D_{\omega}(s,a)}{a \sim \pi_{\theta}(\cdot \mid s)} -\\\expec{D_{\omega}(s,a)}{a \sim \pi_{E}(\cdot \mid s)}
\end{split}
\end{align}
learning to output a high score when encountering pairs produced by $\pi_{E}$, and a low score when conditioned upon outputs from a policy. The output of the critic $D_{\omega}(s,a)$ can then be used as a surrogate reward function $\tilde{r}(s, a)$. Assuming an appropriate value for $\omega$, the surrogate reward increases as actions sampled from $\pi_{\theta}$ look similar to those chosen by experts. In our setting, $\pi_\theta$ may end a training trial prematurely by causing a collision or going off-road. To discourage early stopping, we define $\tilde{r}(s, a)$ to be always positive,
\begin{align}
\label{eq:surrrew}
\tilde{r}(s, a) = \log(1 + e^{D_{\psi}(s,a)})
\end{align}

Optimizing equations~\ref{eq:imit} and ~\ref{eq:wasserstein} has lead to policies that reproduce expert performance in a number of settings \cite{ho16,li2017inferring,kuefler2017imitating}. However, the behavior of these policies tend to be unimodal, failing to account for different latent styles.

\subsection{Information Maximization}

In standard variational information maximization \cite{barber2003algorithm}, the objective is to maximize the mutual information between a generator and posterior $p(z \mid s, a)$ over latent codes by optimizing a lower bound. In contrast, we view $q(z \mid s, a)$ as an inference distribution with associated marginal $q(z)$, rather than a variational approximation to $p(z \mid s, a)$ \cite{zhao2017towards}. We propose maximizing the mutual information between our policy and the joint inference distribution directly, using the factorization in equation~\ref{eq:dbn2}:
\begin{align}
\label{eq:info1}
I_{q}(z ; s, a) = \expec{\log q(z \mid s, a) - \log q(z)}{q(s,a,z,\tau)} \\
\label{eq:info2}
= \expec{\log q(z' \mid s, a)}{\tau, s, z', a} - \expec{\log q(z')}{z'}\\
\label{eq:info3}
= H(Q_{\psi}(z')) - C(\theta, \psi)
\end{align}
where $\tau \sim p(\tau)$ is drawn randomly from a distribution of burn-in demonstrations, the initial observation for the rollout $s \sim p(s \mid \tau)$ is determined by the environment dynamics, and the target latent code $z' \sim q(\cdot \mid \tau)$ and initial action $a \sim \pi_\theta(\cdot \mid s, z')$ must be sampled from learned models.

The model $Q_{\psi}$ is a parametric representation of the inference distribution $q(z \mid s, a)$, parameterized by $\psi$. The objective $C(\theta, \psi)$ is simply the cross entropy error between the latent code $z'$ sampled at the beginning of the trial, and the code predicted by $Q_{\psi}$ at the end, which is minimized in standard InfoGAIL. However, we now sample $z'$ from the inference model $Q_{\psi}(z' \mid \tau)$ conditioned on the burn-in demonstration $\tau$, rather than an arbitrary prior. The term $H(Q_{\psi}(z'))$ is analogous to the entropy over latent codes derived in past work \cite{chen2016infogan,hausman2017multi}. Because we now sample codes from $Q_{\psi}(z \mid \tau)$ at the beginning of each trial, $Q_{\psi}(z') = \expec{Q_{\psi}(z' \mid \tau)}{\tau} \approx \estexpec{Q_{\psi}(z' \mid \tau)}{\tau})$ must be approximated using Monte Carlo estimation.

\subsection{Burn-InfoGAIL}

Combining equations~\ref{eq:wasserstein} and ~\ref{eq:info3}, the final form of our objective is given by:
\begin{equation}
\begin{split}
\min_{\theta} \max_{\omega, \psi} \underbrace{W(\theta, \omega)}_\text{Imitation} - \underbrace{C(\theta, \psi)}_\text{Style} + \lambda \underbrace{H(\estexpec{Q_{\psi}(z' \mid \tau)}{\tau})}_\text{Entropy}
\end{split}
\end{equation}
where $\lambda$ is a hyperparameter controlling the weight of the entropy. The first term encourages the model to imitate the driver data, and the second term allows it to perform its imitation in such a way that the driver class can be predicted from its actions. The third term ensures that the inference model will, on average, sample from among all the latent codes.

Assuming that driver styles are distributed uniformly in the true data set, $H(Q_{\psi}(z'))$ can be interpreted as the Kullback-Leibler (KL) Divergence between the expected value of the inference model and prior distribution. In other words, we sample a code $z$ by conditioning our model on the burn-in demonstration to ensure that the latent code reflects the actual style of the expert each trial. Because the optimization wants to minimize $C(\theta, \psi)$, the sampling posterior may attempt to push its probability mass to a single label, so as to be maximally discriminable. Therefore, $H(Q_{\psi}(z'))$ must be maximized to ensure that, on average, samples from the posterior $Q_{\psi}(z \mid \tau)$ are uniformly distributed. This result leaves open the opportunity to extend our approach to different distributions of expert data by changing the prior over $z$, but we defer this question to future work.

\section{Implementation}

In practice, Burn-InfoGAIL requires an environment simulator in which to generate rollouts and parametric, conditional density estimators to represent the policy, critic, and inference model. This section explains how these components were implemented for our experiments. 

\begin{figure}[!t]
\begin{center}
\includegraphics[width=\columnwidth]{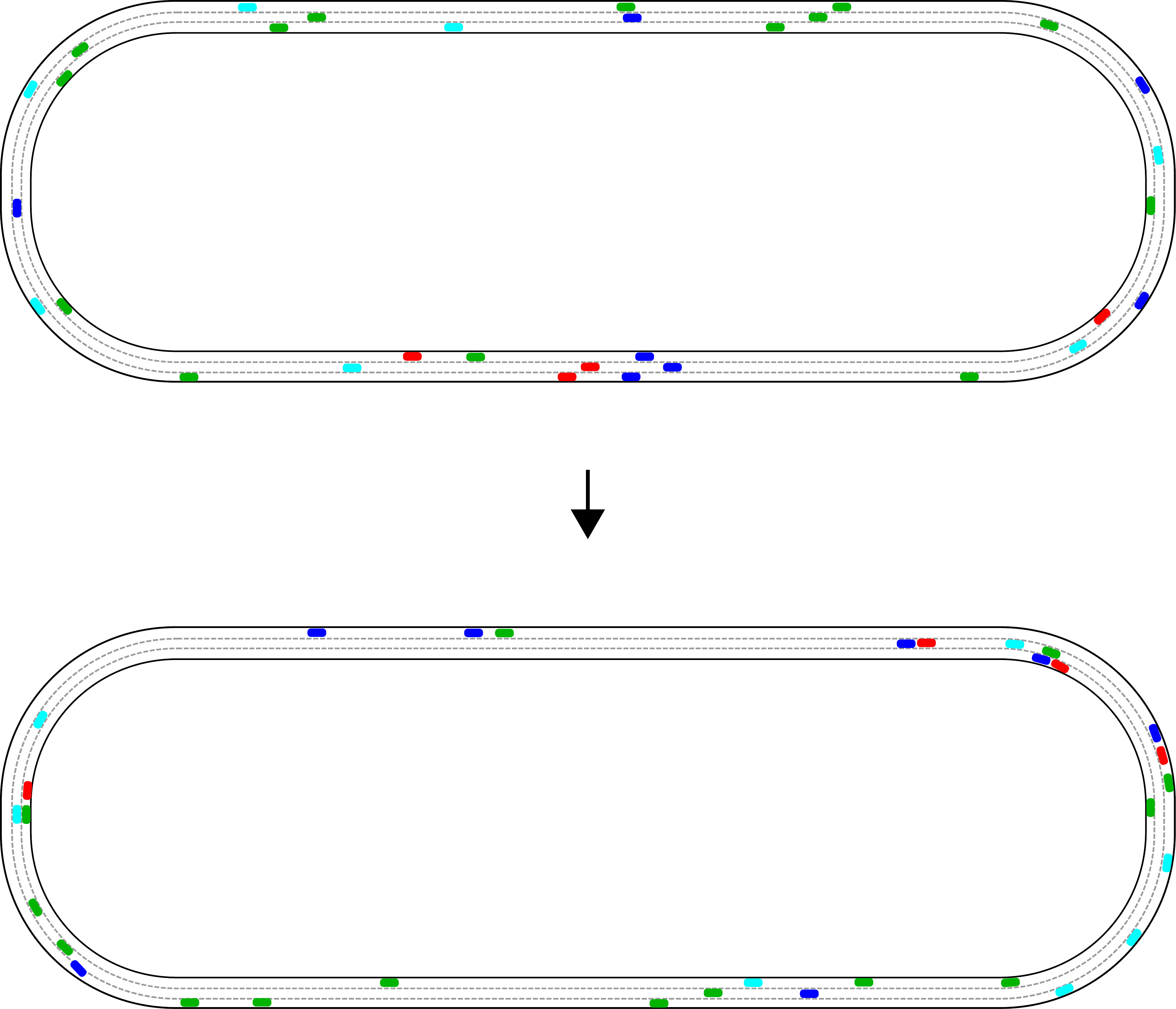}
\caption{Scenes taken from oval track environment after inititialization and a few seconds of driving. Over time, tail-gaiters (green) and aggressive drivers (red) cluster behind passive drivers (blue). Speeders (cyan) retain their large headway distances.}
\label{fig:tracks}
\end{center}
\end{figure}

\subsection{Environment}

The simulator used to generate data and train models is based on an oval racetrack, shown in Figure~\ref{fig:tracks}. As in past work \cite{morton2017simultaneous}, we populate our environment with vehicles simulated by the Intelligent Driver Model \cite{treiber2000congested}, where lane changes are executed by the MOBIL general lane changing model \cite{kesting2007general}. The settings of each controller are drawn from one of four possible parameterizations, defining the style $z$ of each car. The resulting driving experts fall into one of four classes:

\begin{itemize}

\item \textit{Aggressive}: High speed, large acceleration, small headway distances.

\item \textit{Passive}: Low speed, low acceleration, large headway distances.

\item \textit{Speeder}: High speed and acceleration, but large headway distance.

\item \textit{Tailgating}: Low speed and acceleration, but small headway distances.

\end{itemize}

Furthermore, the desired speed of each car is sampled from a Gaussian distribution, ensuring that individual cars belonging to the same class behave differently. A total of 960  training demonstrations and 480 validation demonstrations were used, each lasting 50 timesteps (or 5 seconds, at 10 Hz). 

The observations are represented with a combination of LIDAR and road features \cite{kuefler2017imitating,morton2017simultaneous}. We used 20 LIDAR beams, giving the policy access to both distance and range rate for surrounding cars. Road features included attributes such as the ego vehicle's speed, lane offset, and distance to lane markings. We also include three indicator variables in the observation vector, which detect collision states, offroad events, and driving in reverse. We terminate training when any of these three indicators are activated. Between LIDAR distance, range rate, road features, and indicator variables, the complete observation vector amounts to a total of 51 attributes.

\subsection{Model Architecture}

\begin{figure}[!t]
\begin{center}
\includegraphics[width=0.9\columnwidth]{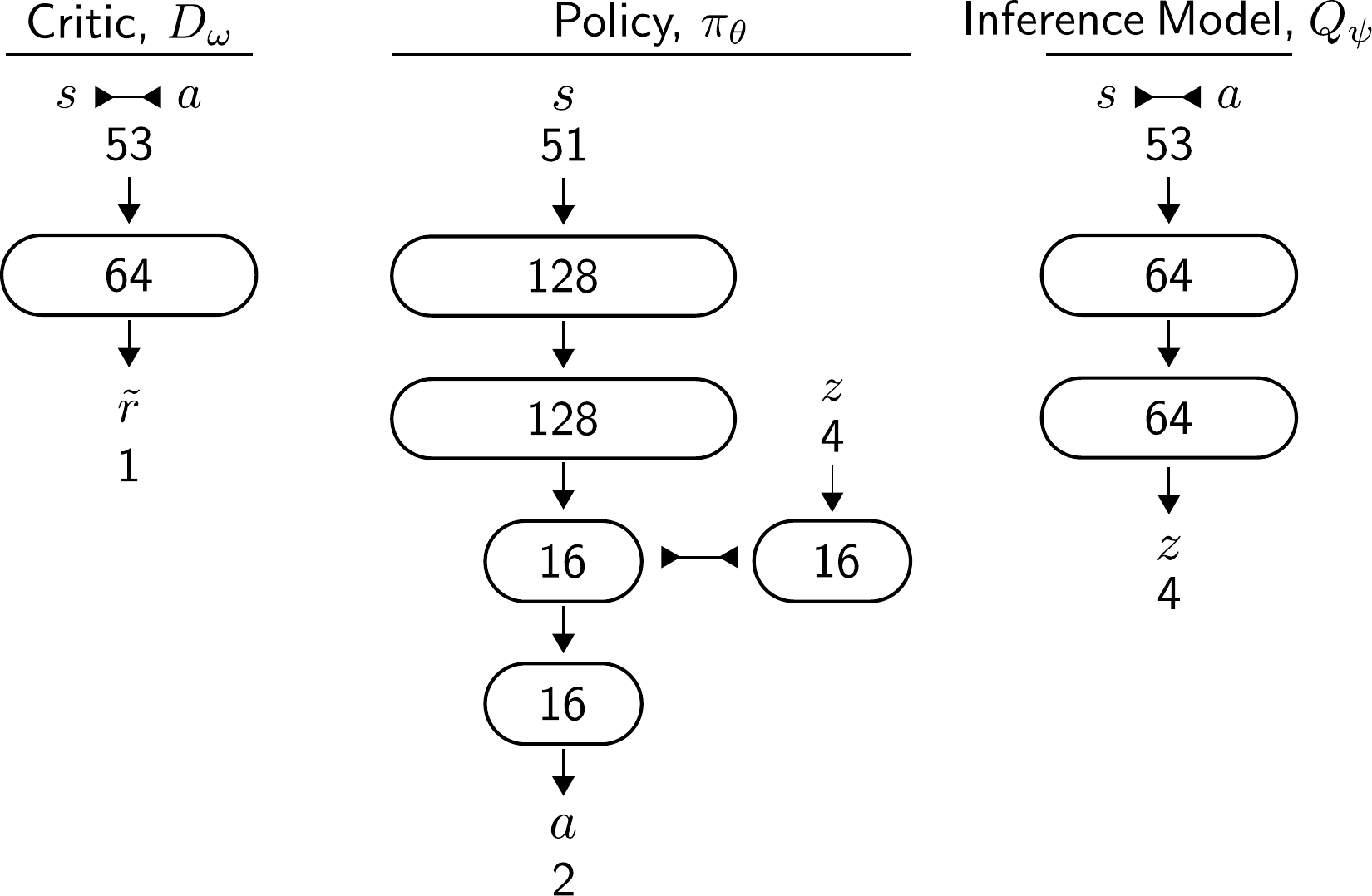}
\caption{Network architecture for the policy $\pi_\theta$, inference model $Q_{\psi}$, and critic $D_\omega$. Directed arrows denote feedforward connections, bidirectional arrows denote concatenation, and integers denote the dimensionality of each layer.}
\label{fig:networks}
\end{center}
\end{figure}

The models $\pi_{\theta}$, $D_{\omega}$, $Q_{\psi}$ (shown in Figure~\ref{fig:networks}) are represented by multilayer perceptrons (MLP) with \textit{tanh} activations. Actions are sampled $a \sim \mathcal{N}(\pi_{\theta}(\cdot \mid s, z), I\sigma)$ during training, where $\sigma$ is also a trainable parameter vector. The 4-dimensional latent code $z$ is passed into $\pi_{\theta}$ using a learned, linear embedding. Because the latent code is of a lower dimensionality than the input features, but we desire it to have a large influence on the outputs of $\pi_{\theta}$, the embedding vector is concatenated with a later hidden layer of the policy network. The policy $\pi_\theta$ attempts to optimize the sum of discounted $\tilde{r}(s,a)$, which is not differentiable with respect to $\theta$. However, policy gradient reinforcement learning can be used to approximate a gradient to train the model iteratively. In this work, we use Trust Region Policy Optimization (TRPO) \cite{schulman2015trust,duan2016benchmarking} to fit $\pi_\theta$.

The inference model $Q_{\psi}$ predicts the parameters of a categorical distribution. Note that although $Q_{\psi}(z \mid s, a)$ is a feedforward network, we condition on trajectories, predicting a value for each state-action pair and taking the most frequent prediction over the sequence. This network is simply trained to perform a $4$-category classification task, where the ``labels'' for each example are generated at the beginning of the trial. Therefore, $Q_{\psi}$ can be trained end-to-end with Adam, which leverages both momentum and feature scaling during stochastic gradient descent \cite{kingma2014adam}.

Finally, the objective used to update $D_{\omega}$ is also differentiable with respect to $\omega$. The class labels (whether a state-action pair was produced by an expert, or $\pi_\theta$) can be determined easily as well. However, \citeauthor{arjovsky2017wasserstein} (\citeyear{arjovsky2017wasserstein}) demonstrate that in order to obey the K-Lipschitz property, momentum free updates must be used to train the discriminator. Therefore, $\omega$ is fit using RMSProp \cite{tieleman2012lecture}.


\section{Experiments}

\begin{figure}[!t]
\begin{center}
\includegraphics[width=\columnwidth]{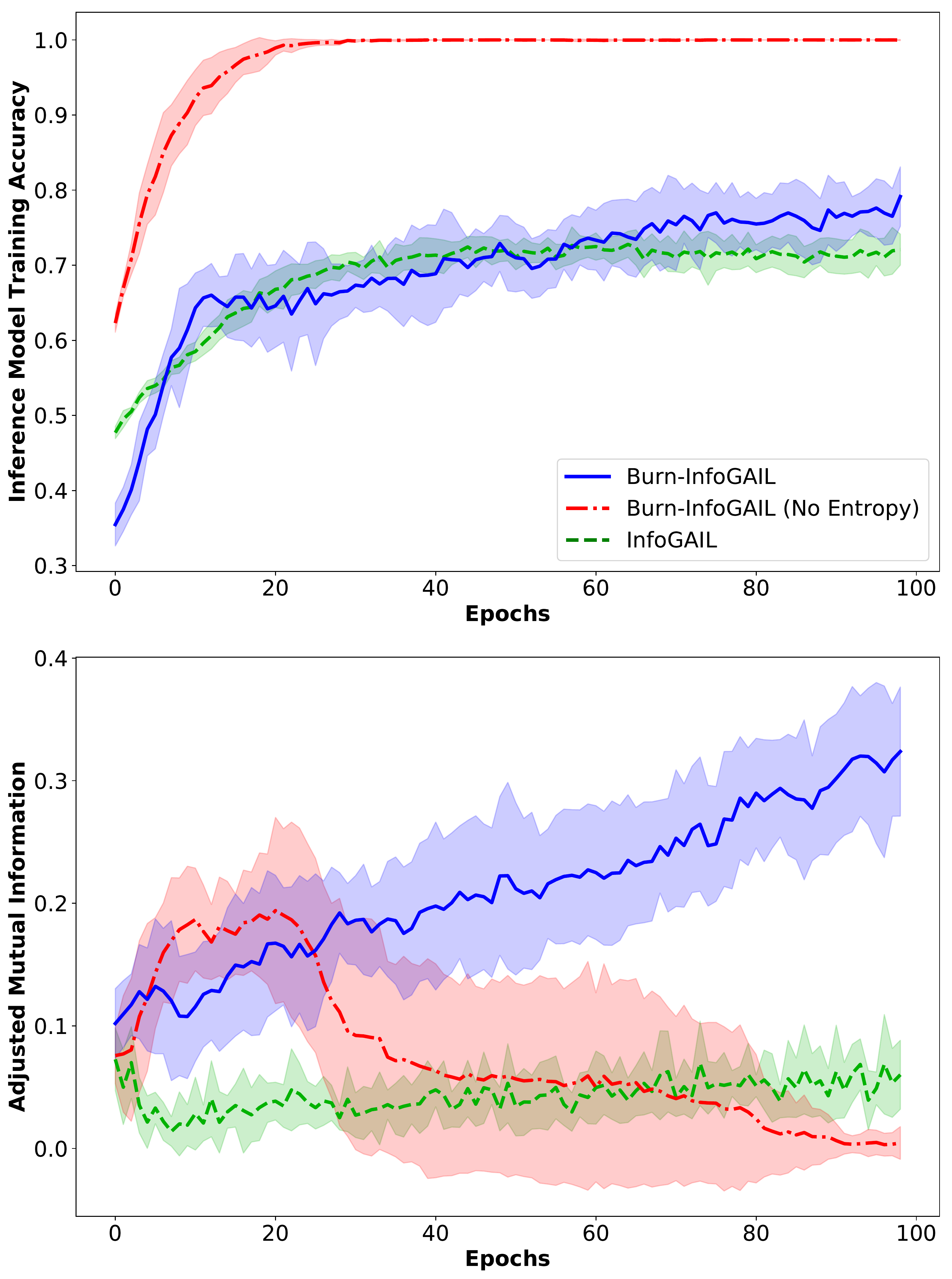}
\caption{Training progress for both the reconstruction accuracy and adjusted mutual information obtained by the inference model, while varying the weight of entropy. Standard deviations were obtained by training 10 models for each experimental condition.}
\label{fig:exp2}
\end{center}
\end{figure}

In the following experiments, we evaluate $\pi_{\theta}$ as a model of driving behavior, and $Q_{\psi}$ as an unsupervised, trajectory clustering technique. We would like to ensure that the values predicted by $Q_{\psi}$, when conditioned on expert trajectories, correlate with the underlying label $z$ of the expert. As such, we use the adjusted mutual information (AMI) to measure performance \cite{vinh2009information}.

\subsection{Entropy and Mutual Information}

\begin{figure}[!t]
\begin{center}
\includegraphics[width=\columnwidth]{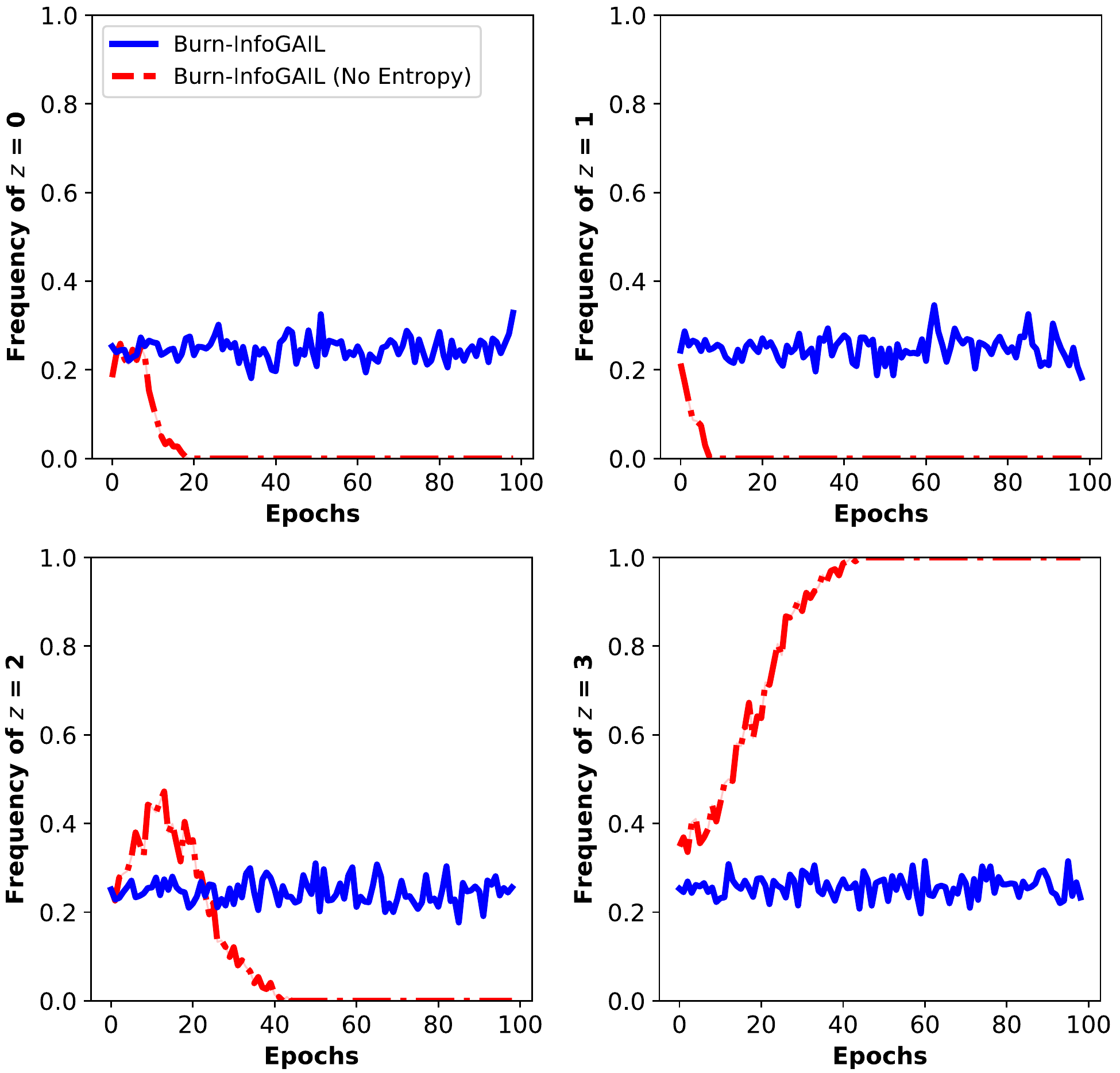}
\caption{The frequency with which each latent code was sampled during model training with different $\lambda$. Entropy weighted models sample the classes uniformly throughout training, whereas models that do not use entropy converge to a single value.}
\label{fig:samp_freq}
\end{center}
\end{figure}

We first experimented with different settings of $\lambda$ in order to assess the role entropy maximization plays in our algorithm. Figure~\ref{fig:exp2} shows that $\lambda = 0$ caused $Q_{\psi}$ to converge to perfect classification accuracy with $\AMI(Q_{\psi}(z \mid \tau),z) = 0$, as predicted. Figure~\ref{fig:samp_freq} gives insight into this degenerate solution. We see that because $Q_{\psi}$ produces its own labels at the beginning of the trial, it learns to collapse the entirety of its probability mass onto a single label (in this case, $Q_{\psi}(z \mid \tau)$ = 3), so as to be maximally predictable. Conversely, when $\lambda = 500$, both AMI and classification accuracy increase over training epochs.

After training, we applied the model achieving the highest AMI to a held out validation set of expert state-action pairs. Table~\ref{tab:baselines} shows that the network outperforms other unsupervised learning techniques on unseen data, including the recurrent, variational autoencoder (VAE) first trained on this task environment \cite{morton2017simultaneous}. 
\begin{table}[]
\centering
\caption{Adjusted mutual information scores of different models of $q(z \mid s, a)$ on validation set. Compares approaches that are unsupervised (U), supervised (S), and those that require a simulator to perform rollouts (R).}
\begin{tabular}{lcc} \toprule
 Method  & Training & Validation AMI   \\ \midrule
 K-Means & U & 0.0 \\
 VAE + K-Means & U & 0.24 \\
 InfoGAIL & U + R & 0.16 \\
 \textbf{Burn-InfoGAIL} & U + R & \textbf{0.38} \\
 SVM & S & 0.95 \\
 VAE + SVM & S & 0.22 \\
 \bottomrule
 
\end{tabular}
\label{tab:baselines}
\end{table}

\subsection{Reproducing Driving Behavior}

Our next experiment tested how policies learned by Burn-InfoGAIL compared to other techniques for imitation learning. We randomly sampled \numrollouts initial conditions and computed the root mean squared error (RMSE) of the speed and global position of learned policies versus expert driving behavior over 30 second trajectories. To ensure that all trajectories had consistent lengths for comparison, the validation environment did not end trials in the event of a collision, offroad, or reversal. Table~\ref{tab:bad_events} shows the frequency with which these ``bad events'' occurred during rollouts for each trial. Burn-InfoGAIL finds a good trade-off between going off-road and avoiding collisions, achieving a collision rate comparable to GAIL, but an off-road rate that is significantly smaller.

We compared against three baseline models: The first baseline is the VAE driver policy proposed by \citeauthor{morton2017simultaneous} (\citeyear{morton2017simultaneous}). Its encoder network consists of two Long Short-Term Memory (LSTM) \cite{hochreiter1997long} layers that map state-action pairs to the mean and standard deviation of a 2-dimensional Gaussian distribution. Its decoder, or policy, is a 2-layer MLP, also consisting of 128 units. During testing, the encoder conditions on the burn-in demonstration and the predicted mean of the distribution is used as the latent code for the policy. The second baseline is a GAIL model trained on the objective in equation~\ref{eq:wasserstein}. It has the same model architecture as $\pi_\theta$, with the exclusion of the learned embedding layer needed to encode the style variable. Finally, we test against an implementation of InfoGAIL that is architecturally identical to $\pi_\theta$, but simply samples $z$ from a discrete uniform distribution at the beginning of each trial.

\begin{table}[]
\centering
\caption{Frequency of dangerous events recorded over \numrollouts rollouts, as a fraction of total timesteps.}
\begin{tabular}{lccc} \toprule
 Method  & Offroad & Collision & Reversal \\ \midrule
 \textbf{Burn-InfoGAIL} & 0.074 & 0.061 & 0.000 \\
 InfoGAIL & 0.033 & 0.099 & 0.126 \\
 GAIL & 0.165 & 0.059 & 0.177 \\
 VAE & 0.756 & 0.021 & 0.000 \\
 \bottomrule
 
\end{tabular}
\label{tab:bad_events}
\end{table}

\begin{figure}[t]
\begin{center}
\includegraphics[width=\columnwidth]{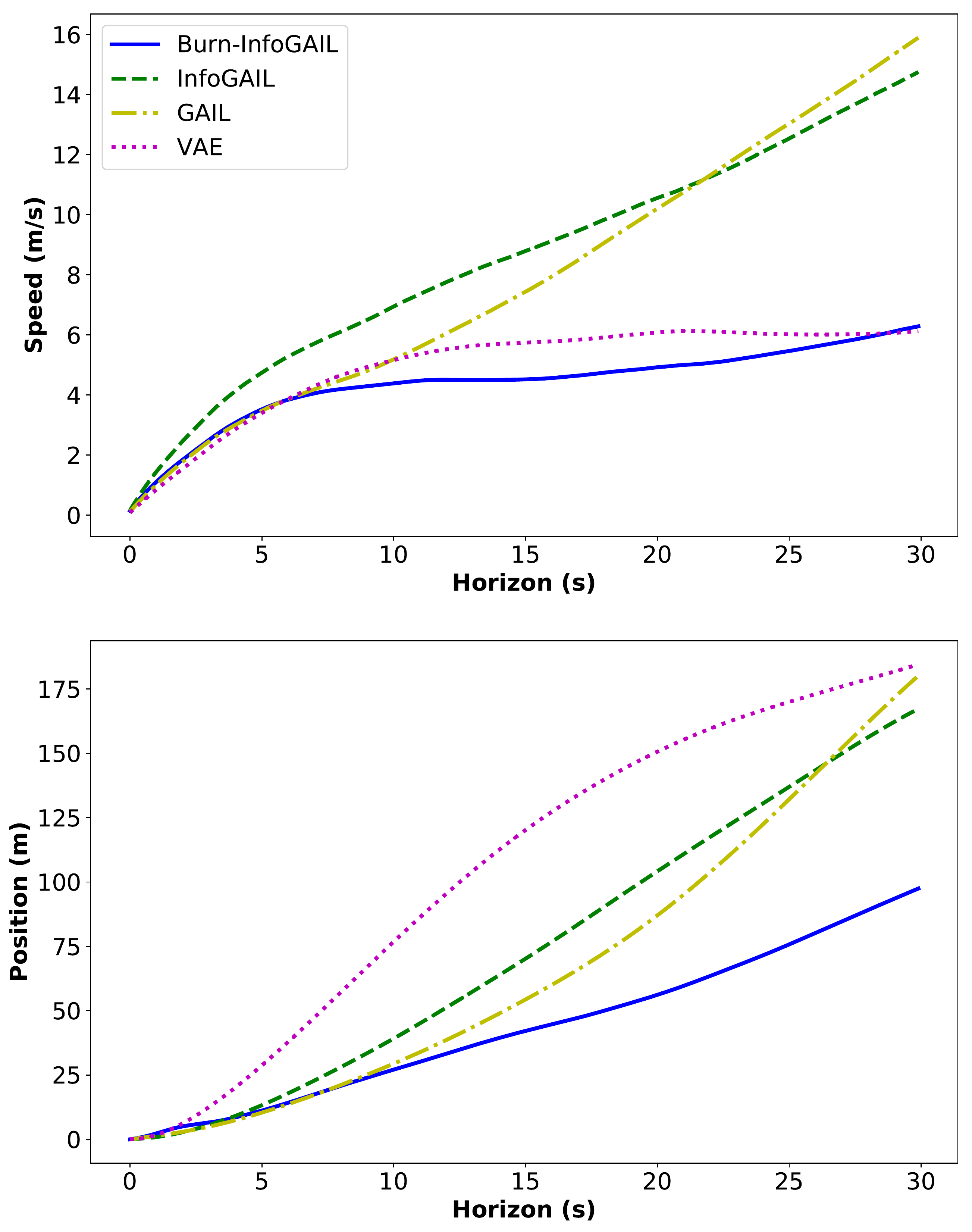}
\caption{Root mean squared error (RMSE) between learned policies and validation trajectories. Results are averaged over \numrollouts rollouts for each model. Our model achieves the lowest error on predicting both speed and position over 30 second trajectories.}
\label{fig:rmse}
\end{center}
\end{figure}

As shown in Figure~\ref{fig:rmse}, Burn-InfoGAIL achieves the lowest error over the longest period of driving. GAIL is able to capture differences in style for about 10 second, presumably because the imitation objective discourages the policy from adjusting its velocity away from its initial conditions. But as minor errors compound over long horizons, GAIL to drifts towards an average policy due to its mode-seeking nature \cite{goodfellow2016nips}. In contrast, the VAE is able to use the latent code inferred from the burn-in demonstration to maintain an appropriate speed, achieving an RMSE close to the true value, rivaling Burn-InfoGAIL. However, being trained without a simulator, the VAE suffers from cascading errors causing it to go off road.

\begin{figure*}[t]
\begin{center}
\includegraphics[width=2\columnwidth]{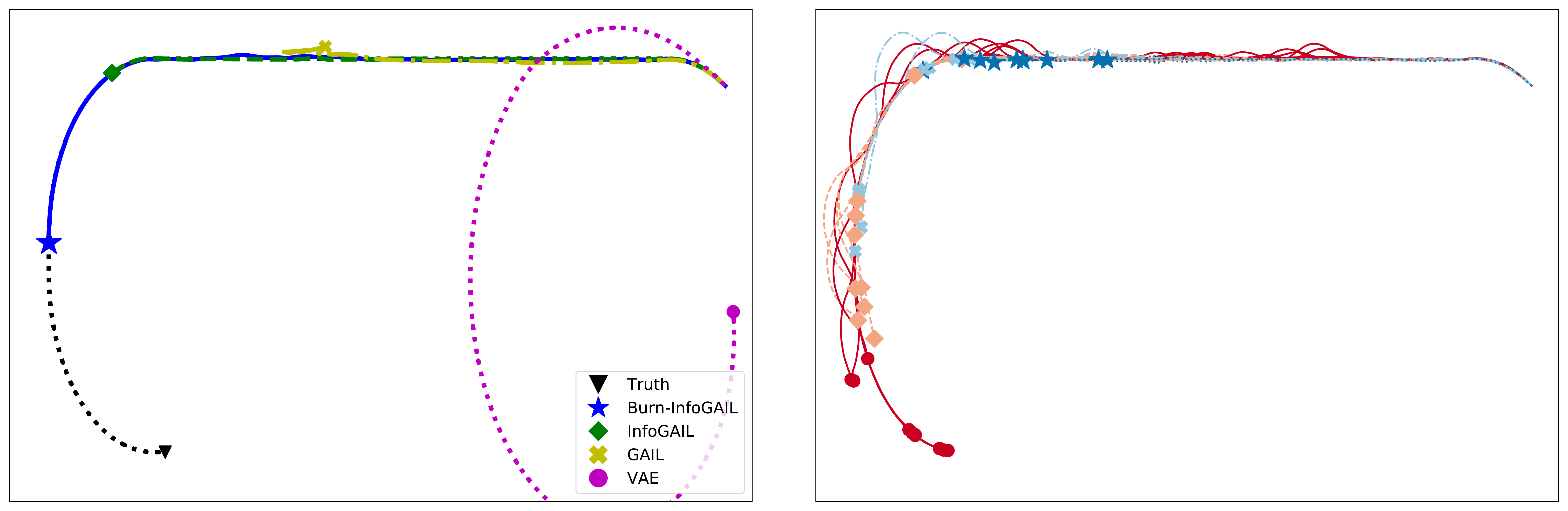}
\caption{Model trajectories on track environment. Left: Example global positions of Burn-InfoGAIL along with baseline models and ground truth ego vehicle. Burn-InfoGAIL tends to end trials closer to the true end point. Right: Driving trajectories subject to sampling different latent codes. We see that terminal states tend to cluster on the basis of the latent code chosen.}
\label{fig:spaghett_plots}
\end{center}
\end{figure*}

\begin{figure}[h]
\begin{center}
\includegraphics[width=\columnwidth]{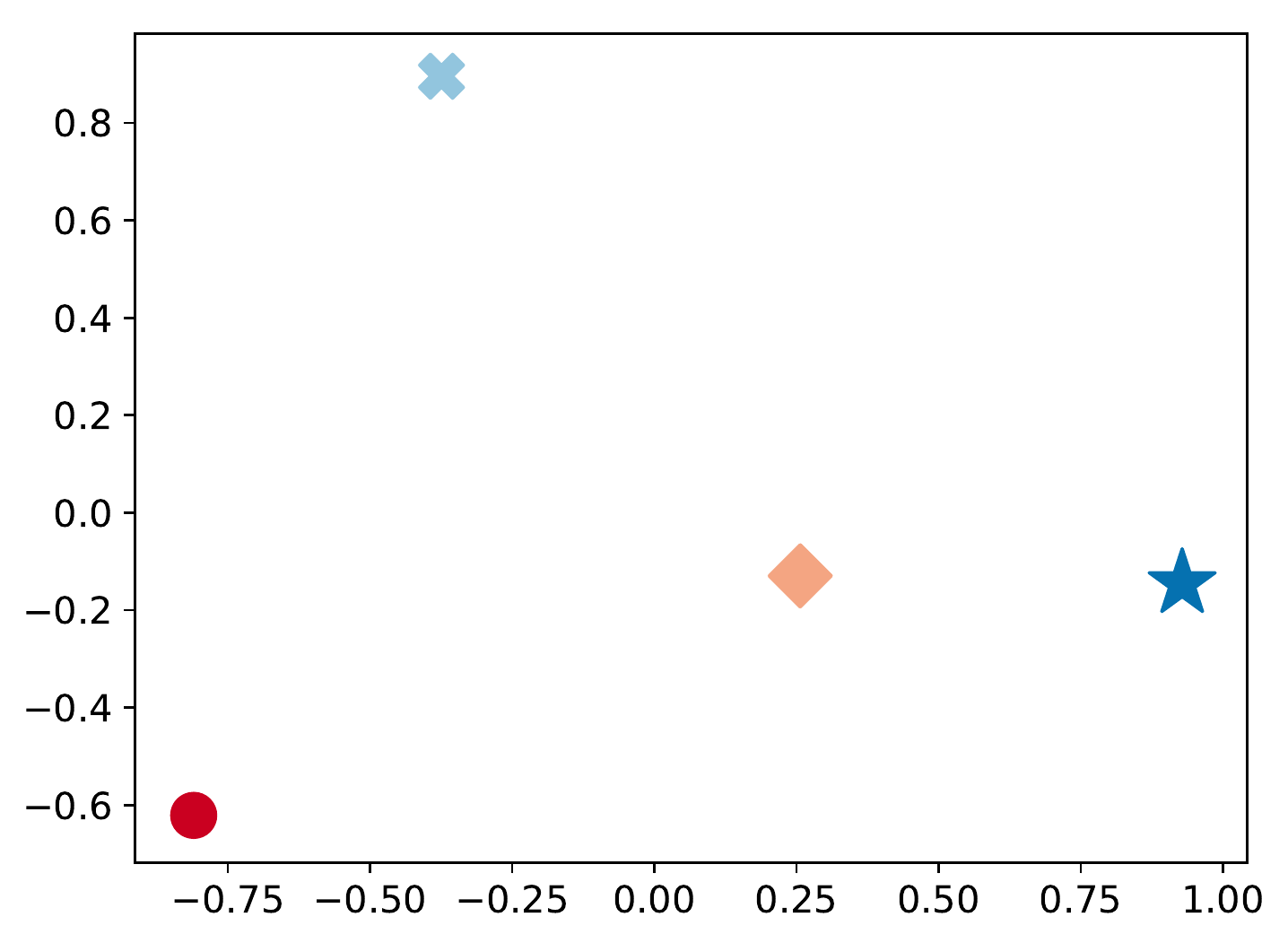}
\caption{Learned embedding vectors (i.e., a subset of $\theta$). Principal components analysis is used to express the 16 dimensional weight vectors in two dimensions. The difference between the most \emph{passive} and \emph{aggressive} driving codes accounts for most of the variance.}
\label{fig:embedding}
\end{center}
\end{figure}

\subsection{Qualitative Results}

Observing that Burn-InfoGAIL obtains low RMSE over many trials, we produced visualizations to assess individual trajectories generated by each model. Figure~\ref{fig:spaghett_plots}a plots the global position of cars driven by each policy (including the IDM expert) over a 30 second period. We see that on the initial straightaway, all models perform comparably. However, the VAE baseline, trained with behavioral cloning, is unable to handle the turn. The GAIL-based techniques follow the curvature of the road more closely, but standard GAIL loses speed over time, ending its trial short of the expert's position. Burn-InfoGAIL, in contrast, maintains the speed of the IDM throughout the drive, finding an endpoint that was closer to the ground truth than the other models.

Starting from the same road scene and ego vehicle, we next sought to understand how sampling different latent codes affected the policy's behavior. Figure~\ref{fig:spaghett_plots}b plots the global position of the car obeying $\pi_{\theta}$. Instead of conditioning our policy on a burn-in demonstration, we select $z$ for $10$ trials, for each possible code. We see that one code (red circle) seems to be designated for aggressive driving, changing lanes more regularly and driving farther (thus achieving a greater velocity) than the other trajectories. In contrast, another code appears to be designated for passive driving (blue star), performing fewer lane changes earlier on and ending closer to the starting position. Like the \textit{speeder} and \textit{tailgater} experts, the other codes tend to fall somewhere in between. When we visualize the learned embedding space of the latent codes by projecting its weight vectors onto two dimensions, we see a similar pattern emerge. Figure~\ref{fig:embedding} demonstrates that most of the variance between the four embeddings is accounted for by the distance between the \emph{aggressive} and \emph{passive} codes, which have the greatest Euclidean distance from one another than the other codes.

\section{Conclusions}

Humans perform many tasks expertly, albeit differently from one another. These differences between expert demonstrations are influenced by latent factors, or underlying styles, that may be determined long before demonstrations are recorded. InfoGAIL successfully extracts the latent factors controlling expert behavior for brief maneuvers \cite{li2017inferring}. Conversely, recurrent VAEs can identify long-term styles but rely on behavioral cloning, and thus produce unstable policies \cite{morton2017simultaneous}. The contribution of this work has been to extend InfoGAIL to control and cluster expert trajectories governed by time invariant styles, as they may exist in sequential decision problems solved by humans. This work also introduced a new formulation of the imitation learning paradigm in which initial states and latent factors are determined by a reference demonstration provided by an expert, and we showed that adopting this formulation along with the Burn-InfoGAIL algorithm leads to realistic models for a simulated, autonomous driving application.

In addressing this problem, we maximize mutual information with respect to a learned, inference distribution rather than maximizing a variational lower bound. We demonstrate that degenerate solutions may be avoided by maximizing the entropy in the estimated marginal distribution over latent codes. Our solution outperforms standard InfoGAIL in clustering time invariant driving styles, outperforming the state of the art on this task environment, while producing driver models that imitate experts over long horizons.

Burn-InfoGAIL appears to produce policies that use their learned, latent code to maintain their velocity over long time horizons. Whereas other GAIL-based approaches regress towards average behavior by the end of their trajectories, Burn-InfoGAIL terminates trials near the end-points of experts. Limitations of the model include its reliance on a simulated rollout environment, limiting its applicability as an unsupervised clustering method. Future work may explore ways to close the reinforcement learning loop, perhaps replacing the full simulation environment with a learned dynamics model for planning and learning from imagined rollouts.

We evaluated our approach on the assumption that expert styles are uniformly distributed, but Burn-InfoGAIL may extend to more uneven distributions. A promising research direction could involve replacing the entropy objective, here used to encourage diversity, with a general KL divergence term between the inference model and a more complex prior over latent codes. This approach may reveal connections between InfoGAIL and hierarchical reinforcement learning paradigms, where the inference distribution intelligently sets tasks as latent codes that the policy diligently follows.

\section{Acknowledgments}

We thank Jayesh Gupta, Jeremy Morton, Rui Shu, Tim Wheeler, and Blake Wulfe for useful discussions and feedback. This material is based upon work supported by the Ford Motor Company.

\balance

\bibliographystyle{aaai}
\bibliography{ref}

\end{document}